\documentclass{article}
\usepackage[accepted]{sysml2019}

\usepackage{microtype}
\usepackage{graphicx}
\usepackage{booktabs} 

\usepackage{hyperref}

\usepackage[english]{babel}
\usepackage{blindtext}
\usepackage{bbm}
\usepackage{amssymb}
\usepackage{amsmath}
\usepackage{pdfpages}
\usepackage{multirow}

\usepackage[T1]{fontenc}

\usepackage{multirow}
\usepackage{graphicx}
\usepackage[table,xcdraw]{xcolor}
\usepackage{rotating}

\usepackage{my_macros}

\makeatletter
\def\thickhline{%
  \noalign{\ifnum0=`}\fi\hrule \@height \thickarrayrulewidth \futurelet
   \reserved@a\@xthickhline}
\def\@xthickhline{\ifx\reserved@a\thickhline
               \vskip\doublerulesep
               \vskip-\thickarrayrulewidth
             \fi
      \ifnum0=`{\fi}}
\makeatother

\newlength{\thickarrayrulewidth}
\usepackage[disable]{todonotes}

\newcommand{\ignore}[1]{}

\newcommand{\cg}{coarse-grained\xspace}
\newcommand{\fg}{fine-grained\xspace}
\newcommand{\cct}{\mbox{CCT}}
\newcommand{\tct}{\mbox{TCT}}

\sysmltitlerunning{\name{}: Resource-Efficient Supervised Anomaly Detection Using Decision Tree-Based Ensemble Methods}

\begin{document}

\twocolumn[
\sysmltitle{\name{}: Resource-Efficient Supervised Anomaly Detection Using Decision Tree-Based Ensemble Methods}




\begin{sysmlauthorlist}
\sysmlauthor{Shay Vargaftik}{vm}
\sysmlauthor{Isaac Keslassy}{vm,technion}
\sysmlauthor{Ariel Orda}{technion}
\sysmlauthor{Yaniv Ben-Itzhak}{vm}
\end{sysmlauthorlist}

\sysmlaffiliation{vm}{VMware Research}
\sysmlaffiliation{technion}{Technion}



\sysmlkeywords{Machine Learning}

\vskip 0.3in

\begin{abstract}

Decision-tree-based ensemble classification methods (DTEMs) are a prevalent tool for supervised anomaly detection. 
However, due to the continued growth of datasets, DTEMs result in increasing drawbacks such as growing memory footprints, longer training times, and slower classification latencies at lower throughput. 
In this paper, we present, design, and evaluate \name{} - a DTEM-based anomaly detection framework that augments standard DTEM classifiers and alleviates these drawbacks by relying on two observations:
(1) we find that a small (\textit{\cg}) DTEM model is sufficient to classify the majority of the classification queries correctly, 
such that a classification is \emph{valid} only if its corresponding confidence level is greater than or equal to a predetermined classification confidence threshold;
(2) we find that in these fewer harder cases where our \cg DTEM model results in insufficient confidence in its classification, we can improve it by forwarding the classification query to one of \textit{expert} DTEM (\textit{fine-grained}) models, which is explicitly trained for that particular case. 
We implement \name{} in Python based on \textit{scikit-learn} and evaluate it over different DTEM methods: RF, XGBoost, AdaBoost, GBDT and LightGBM, and over three publicly available datasets.   
Our evaluation over both a strong AWS EC2 instance and a Raspberry Pi 3 device indicates that \name{} 
offers competitive and often superior anomaly detection capabilities as compared to standard DTEM methods, while significantly improving memory footprint (by up to $5.46\times$), training-time (by up to $17.2\times$), and classification latency (by up to $31.2\times$). 


\end{abstract}
]


\printAffiliationsAndNotice{}





\section{Introduction}
\label{sec:intro}

\subsection{Background and related work}

Supervised anomaly detection includes a wide range of applications such as finance, fraud detection, surveillance, health care, intrusion detection, fault detection in safety-critical systems, and medical diagnosis. 
For example, anomalies in network traffic could mean that a hacked device is sending out sensitive data to an unauthorized destination; anomalies in a credit card transaction could indicate credit card or identity theft; and, anomaly readings from various sensors could signify a faulty behavior in hardware or a software component.
A popular supervised machine learning (ML) solution for anomaly detection is to employ decision-tree-based ensemble classification methods (DTEMs) which rely on either bagging or boosting techniques to improve the detection capabilities, as explained in the following.

\T{Bagging (or bootstrap aggregation)} \cite{breiman1996bagging} is used to reduce the classification variance and by that improve its accuracy. Random Forest (RF) \cite{breiman2001random} is the most well-known decision-tree-based bagging method, which grows each decision tree according to a random subsample of the features and the data instances, resulting in different trees. Then, a majority vote is used to determine the classification.

Several studies \cite{zhang2005network,singh2014big,tavallaee2009detailed,hasan2014support} have proposed using RF for supervised anomaly detection. 
For instance, \cite{zhang2005network} employed RF for anomaly detection by using data mining techniques to select features and handle the class imbalance problem; and, 
\cite{singh2014big} provided a scalable implementation of quasi-real-time intrusion detection system.

RF is a popular classifier as it offers many appealing advantages over other classification methods, such as Neural Networks \cite{ashfaq2017fuzziness}, Support Vector Machines \cite{gan2013anomaly}, Fuzzy Logic methods \cite{bridges2000fuzzy}, and Bayesian Networks \cite{kruegel2003anomaly}. Specifically, RF offers: (1) robustness and moderate sensitivity to hyper-parameters; (2) low training complexity; (3) natural resilience to deal with imbalanced datasets and tiny classes with very little information; (4) embedded feature selection and ranking capabilities; (5) handling missing, categorical and continuous features; (6) interpretability for advanced human analysis for further investigation or whenever such capability is required by regulations \cite{wiki:right_to_explanation}, \eg in order to understand the underlying risks. To that end, an RF can be interpreted by different methods, such as \cite{banerjee2012identifying}. 

All these aforementioned advantages are repeatedly pointed out in the literature via analysis as well as comparative tests (see \cite{resende2018survey,moustafa2018holistic,habeeb2018real} and references therein) especially for intrusion detection (IDS) purposes \cite{zhang2005network,tavallaee2009detailed,hasan2014support}, fraud detection \cite{xuan2018random} and online anomaly detection capabilities \cite{zhao2018online,singh2014big}.



\T{Boosting.} Unlike bagging, boosting primarily reduces classification bias (and also variance). Many popular decision-tree-based boosting methods such as GBDT \cite{friedman2001greedy,hastie2009unsupervised}, XGBoost \cite{chen2016xgboost}, LightGBM \cite{ke2017lightgbm} and AdaBoost \cite{freund1996experiments} employ the boosting concept, usually, by using iterative training. For example, in Adaptive Boosting, a weak classifier such as a \emph{stump} is added at each iteration (unlike bagging methods that use fully grown trees) and typically weighted with respect to its accuracy. Then, the data weights are readjusted such that a higher weight is given to the misclassified instances. In Gradient Boosting, a small decision tree (\eg with 8-32 terminal nodes) is added at each iteration and scaled by a constant factor. Then, a new tree is grown to reduce the loss function of the previous trees. For both methods, the next trees are trained with more focus on previous misclassifications.

Decision-tree-based boosting methods are known to be among the best off-the-shelf supervised learning methods available \cite{roe2006boosted,schapire2003boosting,liu2017visual,roe2005boosted}, achieving excellent accuracy with only modest memory footprint, as opposed to RF that is usually memory bounded. Boosting methods also share many of the aforementioned advantages offered by RF such as natural resilience to deal with imbalanced datasets and tiny classes, embedded feature selection, and ranking capabilities, and handling missing, categorical and continuous features \cite{medium2017boosing}.

Decision-tree-based boosting methods are also known to be especially appealing for anomaly detection purposes where data is often highly imbalanced (\eg credit card transactions or cyber-security) \cite{pfahringer2000winning,li2012robust}. This is mainly because decision-tree-based boosting methods alter their focus between the different iterations on the more difficult training instances. This often produces a stronger strategy to deal with imbalanced datasets by strengthening the impact of the anomalies and, when adequately trained, boosting methods may usually achieve higher accuracy (as well as precision and recall) than a traditional RF classifier. 

That being said, boosting methods are also more sensitive to overfitting than RF, especially when the data is noisy \cite{dietterich2000experimental}. The training of boosting-based methods generally takes much longer than RF, mainly since the trees are built sequentially and compute-intensive tasks such as classification and data weights readjustments take place at every iteration. Moreover, boosting-based methods are harder to tune as compared to RF, as they have more parameters and higher sensitivity to these parameters. 

Finally, both bagging (RF) and boosting methods are prevalent tools for supervised anomaly detection with shared and distinct pros and cons, and there is no clear winner in this classification contest as the best classifier often depends on the specific dataset and the application.

\subsection{Challenges} 

In recent years, supervised anomaly detection via DTEMs is becoming extremely difficult. This is because traditional bagging DTEMs (\ie random forest) classifiers can be highly effective, but tend to be memory bound, and slower at classification \cite{liaw2002classification,van2012accelerating,mishina2015boosted}. Furthermore, the classification latency of an RF increases with the RF depth \cite{asadi2014runtime}. Accordingly, previous work suggested different approaches to tackle the memory and performance drawbacks of RF. The study of \cite{van2012accelerating} achieves deterministic latency by producing compact random forests composed of many, small trees rather than fewer, deep trees. 
Other studies \cite{asadi2014runtime,browne2018forest} optimize memory-layouts of RF, which reduces cache misses. 

On the other hand, boosting DTEMs are slow to train and tune and also admit slower classification as the number of trees increases \cite{appel2013quickly,rfvsboostblog}. Much effort has been made to address these drawbacks. For example, recent scalable implementations of the tree-based gradient boosting methods include: XGBoost \cite{chen2016xgboost} that supports parallelism and uses pre-sorted and histogram-based algorithms for computing the best split; LightGBM \cite{ke2017lightgbm} that uses a novel technique of Gradient-based One-Side Sampling (GOSS) to filter out the data instances for finding a split value; CatBoost \cite{prokhorenkova2018catboost} that implements \emph{ordered boosting}, a permutation-driven alternative to the classic algorithm and an innovative algorithm for processing categorical features (\eg giving indices of categorical columns such that it can be encoded by one-hot encoding). For Adaptive boosting (\ie AdaBoost), several approaches have been suggested as well to accelerate its slow training \cite{chu2004fast,seyedhosseini2011fast,olson2017jousboost}. For example \cite{seyedhosseini2011fast} introduces a new sampling strategy (WNS) that selects a representative subset of the data at each iteration and by that reducing the number of data points onto which AdaBoost is applied. 

Nevertheless, both bagging and boosting DTEM methods are challenged by the continuous growth of datasets \cite{li2014scaling} in terms of the number of features, data instances, and the increasing demand for lower memory footprints, faster training, and lower classification latency. 
That is, while a sufficiently large DTEM classifier may offer a satisfactory anomaly detection capabilities when adequately trained, it would typically suffer from at least one of the following drawbacks: (1) large memory footprint; (2) long training (which also incurs high energy consumption); (3) high classification latency and low classification throughput. 

Overcoming these drawbacks is essential for efficient DTEM-based anomaly detection systems that are required to have the ability to quickly train a proper classifier in a timely manner and to offer low classification latency and high throughput at reasonable memory footprints and costs.
 
\subsection{Contributions} 

In this paper, we present \name{}, which addresses the aforementioned drawbacks of DTEM methods. \name{} is orthogonal to the discussed bagging and boosting techniques and can augment them to form more efficient DTEM classifiers. We implement and evaluate \name{} for RF, XGBoost, AdaBoost, GBDT, and LightGBM. 

The design of \name{} mainly relies on the two following observations, which are further detailed in Section~\ref{sec:Preliminaries}: 

(1) We find that a small (\textit{\cg}) DTEM model is sufficient to classify the majority of the classification queries correctly. To that end, we define a confidence level threshold, such that a classification is considered to be \textit{valid} only when its classification confidence level is higher than or equal to this given threshold.  

(2) We find that in these fewer harder cases where our \cg DTEM model exhibits insufficient confidence in its classification, we can improve it by forwarding the classification query to one of expert DTEM (\textit{fine-grained}) models, which is explicitly trained for that particular case. 

Finally, in Section~\ref{sec:evaluation}, we present evaluation results over three publicly available datasets and on a strong AWS EC2 instance as well as on a Raspberry Pi 3 device. The results are consistent and indicate that \name{} always offers competitive and often superior anomaly detection capabilities as compared to the standard DTEM methods. For Bagging (RF), \name{} significantly improves the training-time (by up to $6.42\times$), classification latency (by up to $7.23\times$), and memory consumption (by up to $5.46\times$); For Boosting (\eg XGBoost and AdaBoost), \name{} significantly improves the classification latency (by up to $31.2\times$) and training time (by up to $17.2\times$), while being competitive in model memory footprint (always in the range of [0.64$\times$, 2.92$\times$]).


\begin{figure}[!t]
	\centering
	\begin{subfigure}{0.49\linewidth}
		\includegraphics[width=\textwidth]{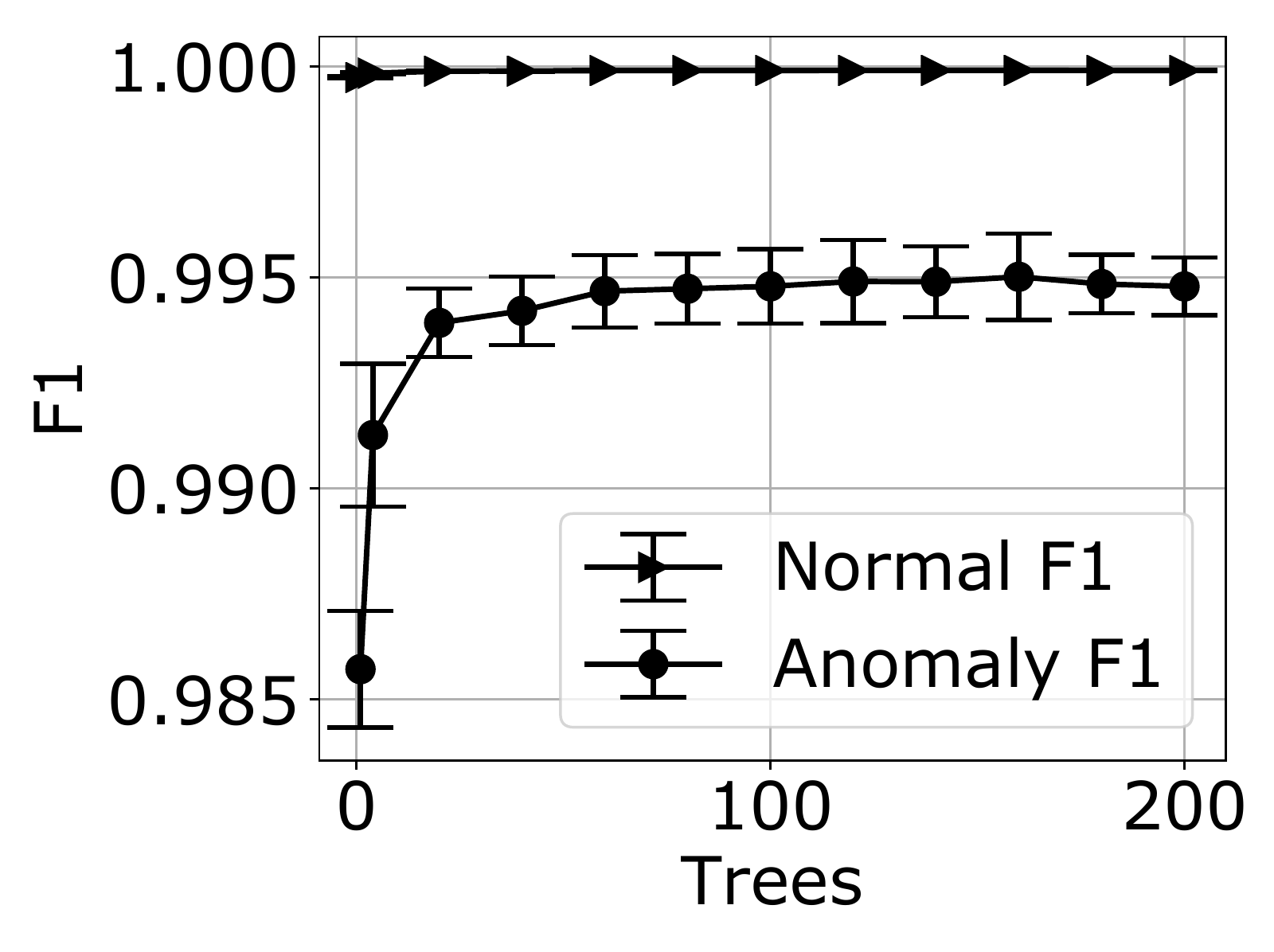}
		\caption{RF over KDD.}
		\label{fig:baseline_rf_kdd}
	\end{subfigure}
	\begin{subfigure}{0.49\linewidth}
		\includegraphics[width=\textwidth]{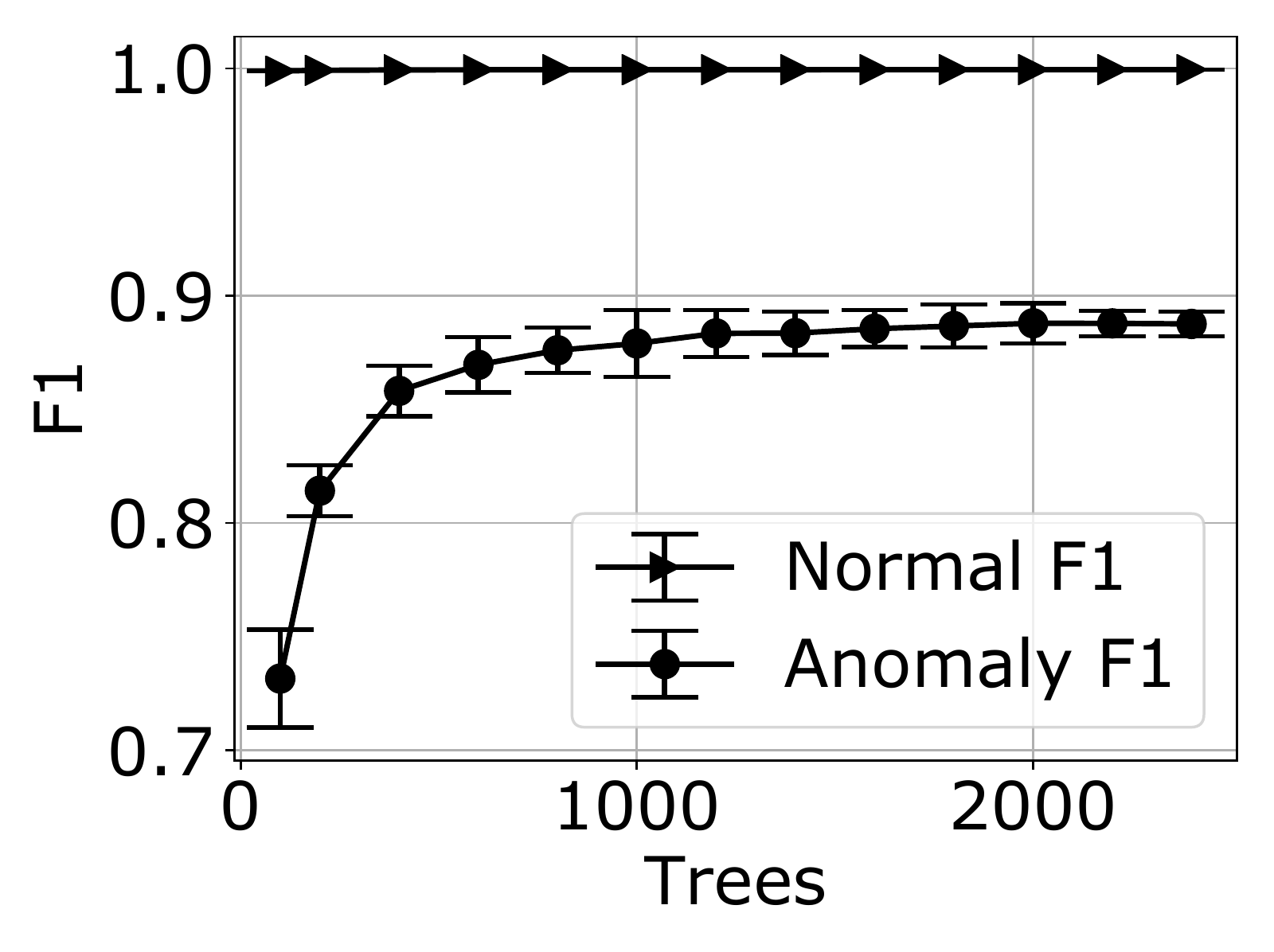}
		\caption{XGBoost over FC.}
		\label{fig:baseline_xgboost_forestcover}
	\end{subfigure}
    \vspace{-6pt}
	\caption{Baseline tuning examples by sweeping over the number of trees. (a) The best $F_1$ score for the RF classifier over the KDD dataset is achieved for 150 trees with no depth limitation, \ie $C(150, None)$. (b) The best $F_1$ score for XGBoost classifier over the FC dataset is achieved for 2000 trees with depth limitation of 3, \ie $C(2000,3)$.}
	\vspace{-6pt}
	\label{fig:baseline_tuning_examples}
\end{figure}

\begin{table}[!t]
	\centering	
	\caption{Baseline DTEM models. Number of trees and tree depth limitations for each DTEM classifier and dataset (RF is trained without tree depth limitation).}
	\vspace{-6pt}
	\label{tab:baseline_sizes}
	\resizebox{0.99\linewidth}{!}{%
		\begin{tabular}{c|c|c|c|}
			\cline{2-4}
			& KDD & CCF & FC \\ \hline
			\multicolumn{1}{|c|}{RF} & $C(150,None)$ & $C(85,None)$ & $C(80,None)$ \\ \hline
			\multicolumn{1}{|c|}{GBDT} & $C(1400,5)$ & $C(600,5)$ & $C(800,5)$ \\ \hline
			\multicolumn{1}{|c|}{XGBoost} & $C(450,3)$ & $C(500,3)$ & $C(2000,3)$ \\ \hline
			\multicolumn{1}{|c|}{LightGBM} & $C(1000,5)$ & $C(1600,5)$ & $C(2800,3)$ \\ \hline
			\multicolumn{1}{|c|}{AdaBoost} & $C(700,3)$ & $C(1100,2)$ & $C(1000,3)$ \\ \hline
		\end{tabular}%
	}
    \vspace{-6pt}
\end{table}


\begin{figure*}[t!]
    \centering
    \includegraphics[width=\linewidth,trim={0cm 0cm 0cm 2.3cm},clip]{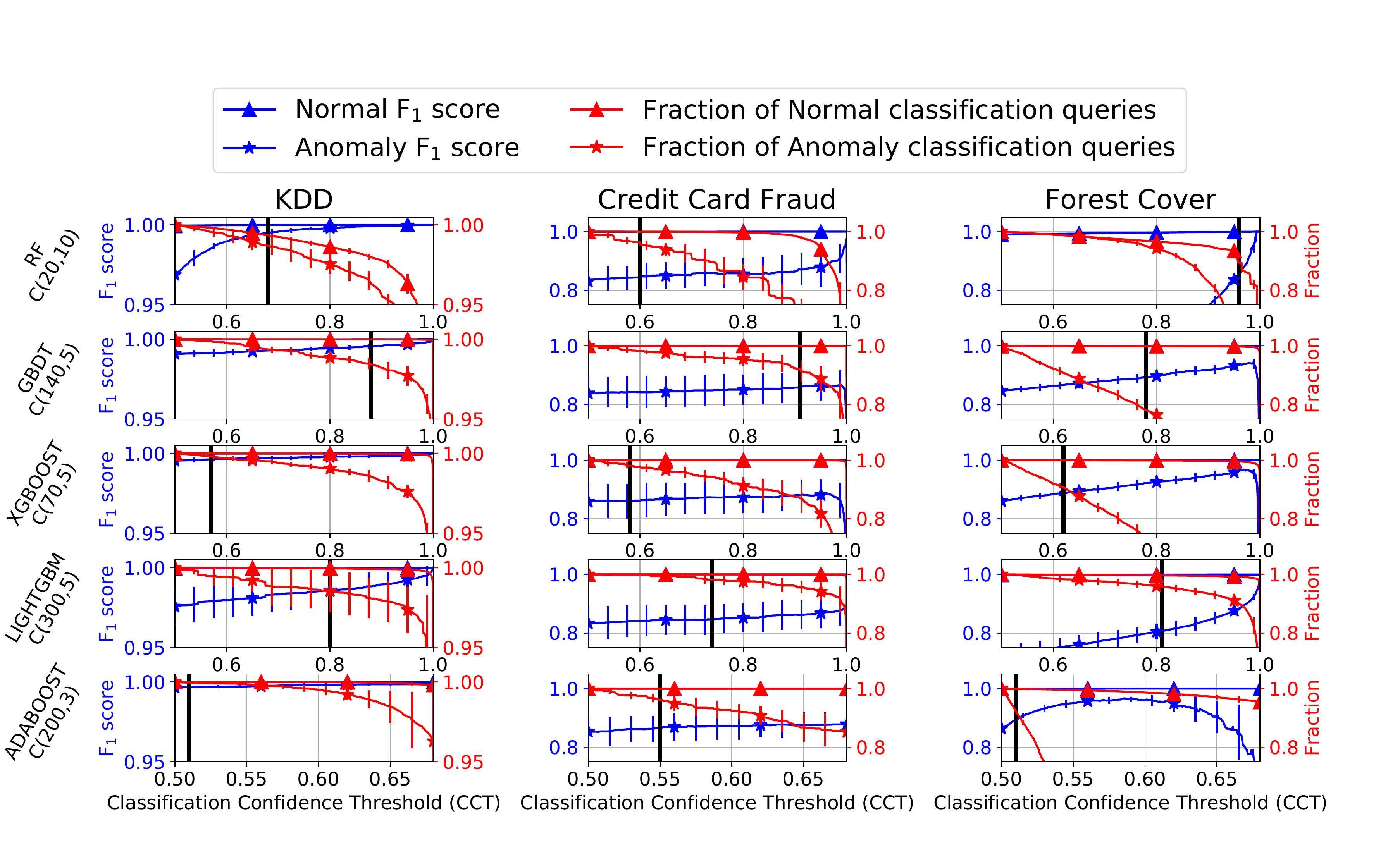}
    \vspace{-30pt}
    \caption{The useful classification fraction and its resulting $F_1$ score by a small (\cg) model, when a classification is \textit{valid} only if its confidence level is greater than or equal to a given classification confidence threshold (\cct). Our observation is that most classifications can be achieved by a \cg model, with a similar/higher resulting $F_1$ score as compared to the $F_1$ score achieved by a much bigger model -- termed baseline model\textsuperscript{\ref{footnote}}.
    The vertical lines indicate the lowest \cct{} for each dataset and model such that the resulting $F_1$ score exceeds the $F_1$ score by a baseline model (see Table \ref{tab:baseline_sizes} for details). Note that any other \cct{} value can be set according to the desired tradeoff between the useful classification fraction and its resulting $F_1$ score. Demonstrated for KDD \cite{kdd_dataset}, Credit Card Fraud (CCF) \cite{ccf_dataset}, and Forest Cover (FC) \cite{forest_cover} datasets, over five different DTEM classifiers, RF \cite{breiman2001random}, GBDT \cite{friedman2001greedy,hastie2009unsupervised}, XGBoost \cite{chen2016xgboost}, LightGBM \cite{ke2017lightgbm} and AdaBoost \cite{freund1996experiments}. In all evaluations, we use a 5-fold cross validation and depict the mean value and variance.}
    \vspace{-12pt}
    \label{fig:coarse_grained}
\end{figure*}


\section{\name Preliminaries}
\label{sec:Preliminaries}

\subsection{Baseline DTEM models}

We evaluate \name{} and quantify its benefits by comparing it to standard DTEM models we term \emph{baseline} models. Specifically, we define a baseline model for three different datasets (KDD \cite{kdd_dataset}, Credit-Card Fraud (CCF) \cite{ccf_dataset}, and Forest Cover (FC) \cite{forest_cover}) and for five different DTEM classifiers (RF \cite{breiman2001random}, GBDT \cite{friedman2001greedy,hastie2009unsupervised}, XGBoost \cite{chen2016xgboost}, LightGBM \cite{ke2017lightgbm} and AdaBoost \cite{freund1996experiments}).
For each dataset and DTEM classifier, we choose the baseline model for comparison by sweeping over different parameters and conducting 5-fold cross-validation for each measurement. Figure \ref{fig:baseline_tuning_examples} shows two such examples. In these specific examples, we depict the $F_1$ score as a function of the number of trees for the RF model over the KDD dataset (\ref{fig:baseline_rf_kdd}) and for the XGBoost model over the FC dataset (\ref{fig:baseline_xgboost_forestcover}). Note that we conduct sweeps over the number of trees for different parameters (\eg class/sample weights for RF, the number of features to consider for a split, tree depth limitations, learning rate for XGBoost) and in these examples, the other parameters are already chosen accordingly.

\T{Remark.} Interestingly, these two sweeps already point out the weaknesses of bagging and boosting methods: (1) The sweep in Figure \ref{fig:baseline_rf_kdd} lasted for 699 seconds and required 205~MB, whereas a similar sweep for XGBoost (over KDD) lasted as much as 14865 seconds but required only 8~MB; (2) Similarly, the sweep in Figure \ref{fig:baseline_xgboost_forestcover} lasted for 25442 seconds and required 47~MB, whereas a similar sweep for RF (over FC) lasted only 425 seconds but required 340~MB\footnote{These specific baseline tuning examples were conducted on an Intel(R) Core(TM) i7-7700 CPU @ 3.60GHz with 4 Cores and 8 Logical Processors.}.

\T{ML performance metric.} In this paper, we use \emph{per-class} $F_1$ score to quantify the anomaly detection capabilities of ML models since this score takes into account the imbalanced nature of the datasets in anomaly detection use-cases. 
Nevertheless, we also consider the Area Under the Curve (AUC) and Average Precision (which is more suitable for skewed datasets) metrics, and obtain similar results.  

\T{Classifier configuration.} For ease of exposition, we denote a classifier with $T$ trees, each limited to a depth of $D$, by $C(T,D)$. A classifier without a depth limitation is denoted by $C(T,None)$.
Table \ref{tab:baseline_sizes} summaries the baseline DTEM classifiers which we use throughout the paper.

\subsection{Observations}
\label{subsec:observations}

In this work, we target binary supervised anomaly-detection classification with \textit{Normal} and \textit{Anomaly} imbalanced classes, and discuss multi-class anomaly detection as a future direction/extension of \name{} in Section \ref{sec:future-work}. In the following, we describe the two main aforementioned observations which our solution is based on.

\T{Observation 1: A small DTEM model can classify most of the classification queries with a high $F_1$ score.}

Figure \ref{fig:coarse_grained} exemplifies via three datasets (KDD, CCF and FC. Further details are in Section~\ref{subsec:datasets}), and over five DTEMs (RF, GBDT, XGBoost, LightGBM, and AdaBoost) that a small DTEM model, which we term \textit{\cg} model, can be used to correctly classify the majority of classification queries (but not necessarily all) by requiring a sufficiently high classification confidence level\footnote{For a DTEM classifier, the classification and its confidence level is determined by the classification distribution vector. For example, if the classification output of an instance is (Normal=0.78, Anomaly=0.22) that means the instance is classified as Normal with a classification confidence level of 0.78.}.
That is, the classification result of the \cg model is \emph{valid} only if its corresponding confidence level is greater than or equal to a predetermined classification confidence threshold (denoted by \cct), rather than simply accepting any classification confidence\footnote{For example, if the prediction output of a sample is (Normal=0.83, Anomaly=0.17) and \cct$=0.9$ that means the classification is not valid and this query requires further attention as we later detail in Section \ref{sec:system-label}.}. We empirically find that this approach of setting a higher \cct{} value to make a valid classification, results in a high fraction of the data instances being valid classifications with a high $F_1$ score for both Normal and Anomaly classes. 

Specifically, as can be seen in Figure \ref{fig:coarse_grained}, the fraction of the valid classifications (out of the total data instances) reduces as \cct{}  increases, and their respective $F_1$ score increases\footnote{Unless the \cct{} is set too high, such that the fraction of valid classifications, especially anomalies, drops to nearly zero and the remaining few instances may admit a high classification variance (See XGBoost over CCF in Figure \ref{fig:coarse_grained} for example). Such \cct{} values are not reasonable operating points for \name{}.\label{footnote}} since only the classifications with a higher confidence level (\ie valid) are being considered\footnote{Our usage of a classification confidence level is inherently different from the threshold used to produce a ROC curve. That is, when producing a ROC curve, the threshold only determines the result of a classification query and not whether it is valid.}. 
Furthermore, the fraction of the Normal data instances is significantly higher than the Anomaly fraction as \cct{} increases. Intuitively, this is because the Anomaly labeled instances are harder to classify, as there are significantly fewer Anomaly instances than the Normal instances available for training.

We have found this observation to be consistent for all tested datasets. These include the three datasets discussed in this paper, as well as a Bankruptcy data set \cite{url:bankruptcy,zikeba2016ensemble}, a Shuttle data set \cite{shuttle_dataset} and different synthetic datasets generated by the scikit-learn machine learning Python package \cite{scikit-learn}.

As we later discuss and demonstrate, any \cct{} value can be set according to the desired tradeoff between the useful classification fraction and its resulting $F_1$ score. 
The vertical black lines in Figure \ref{fig:coarse_grained} indicate the lowest \cct{} for each dataset and classifier such that the resulting $F_1$ score of the valid classifications of both the Normal and Anomaly classes exceed the total $F_1$ score of its corresponding baseline model, as detailed in Table \ref{tab:baseline_sizes} (for the baseline models, a valid classification is any classification with a confidence level of a least 0.5, \ie all classification queries are valid and considered). Finally, we provide two examples to clarify the findings depicted by Figure \ref{fig:coarse_grained}.

\textit{Example 1.} By setting \cct${=}0.995$ (instead of the usual $0.5$) when using the RF model, $C(20,10)$,  for the CCF dataset, we obtain that a fraction that accounts for $51.6\%$ of the Normal data instances is classified with $F_1$ score of $\approx 0.99991$; and a fraction that accounts for  $45.5\%$ of the Anomaly data instances is classified with $F_1$ score of $\approx 0.943$. Whereas, the corresponding baseline DTEM model, $C(85,None)$, achieves $F_1$ score of $\approx 0.99974$ and $\approx 0.8344$, respectively. 

\textit{Example 2.} By setting \cct${=}0.99$ (instead of the usual $0.5$) when using the XGBoost model, $C(70,5)$, for the KDD dataset, we obtain that a fraction that accounts for $99.8\%$ of the Normal data instances is classified with $F_1$ score of $\approx 0.99993$; and a fraction that accounts for  $94.2\%$ of the Anomaly data instances is classified with $F_1$ score of $\approx 0.9992$. Whereas, the corresponding baseline DTEM model, $C(450,3)$, achieves $F_1$ score of $\approx 0.99993$ and $\approx 0.9962$, respectively. 

\T{Observation 2: Train expert (\fg) classifiers to succeed specifically where the \cg model is not sufficiently confident.}
 
When applying the approach suggested by the previous Observation 1, we remain with a small fraction of the data instances without a valid classification by the \cg model due to an insufficient classification confidence level (\ie queries for which the top-1 class probability is lower than \cct). These queries are the harder data instances and, most importantly, as depicted in Figure \ref{fig:coarse_grained}, contain most of the anomalies as the \cct{} increases. 

Our second main observation is that we can leverage the classification distribution vector of the \cg model over the \emph{training data} to: (1) \emph{filter most of the training data} by using a training confidence threshold (\tct), and to (2) train expert classifiers, which we term \textit{\fg} models, that are \emph{trained to succeed specifically where the \cg model is not sufficiently confident and is more likely to make a classification mistake}. The training dataset of each \fg classifier is defined according to \tct{} and the resulting classification distribution vectors of the \cg model (see Section \ref{sec:system-label} for more details).

As we show in Section~\ref{subsec:tuning}, these training datasets of the \fg classifiers are tailored such that they focus on the harder data instances and improve the Normal to Anomaly ratio of the labeled instances as compared to the training dataset of the \cg and baseline classifiers. As a result, we find that the \fg models achieve better $F_1$ score for the low-confidence data instances. 

Furthermore, these tailored training datasets are much smaller as compared to the training sets of the \cg and baseline classifiers, which in turn significantly reduces the required training time (and hence the corresponding energy consumption). This attribute is especially appealing for the boosting methods (see Section \ref{sec:evaluation}).


\begin{figure}[t!]
	\centering
	
	\begin{subfigure}{0.85\linewidth}
		\includegraphics[width=\textwidth]{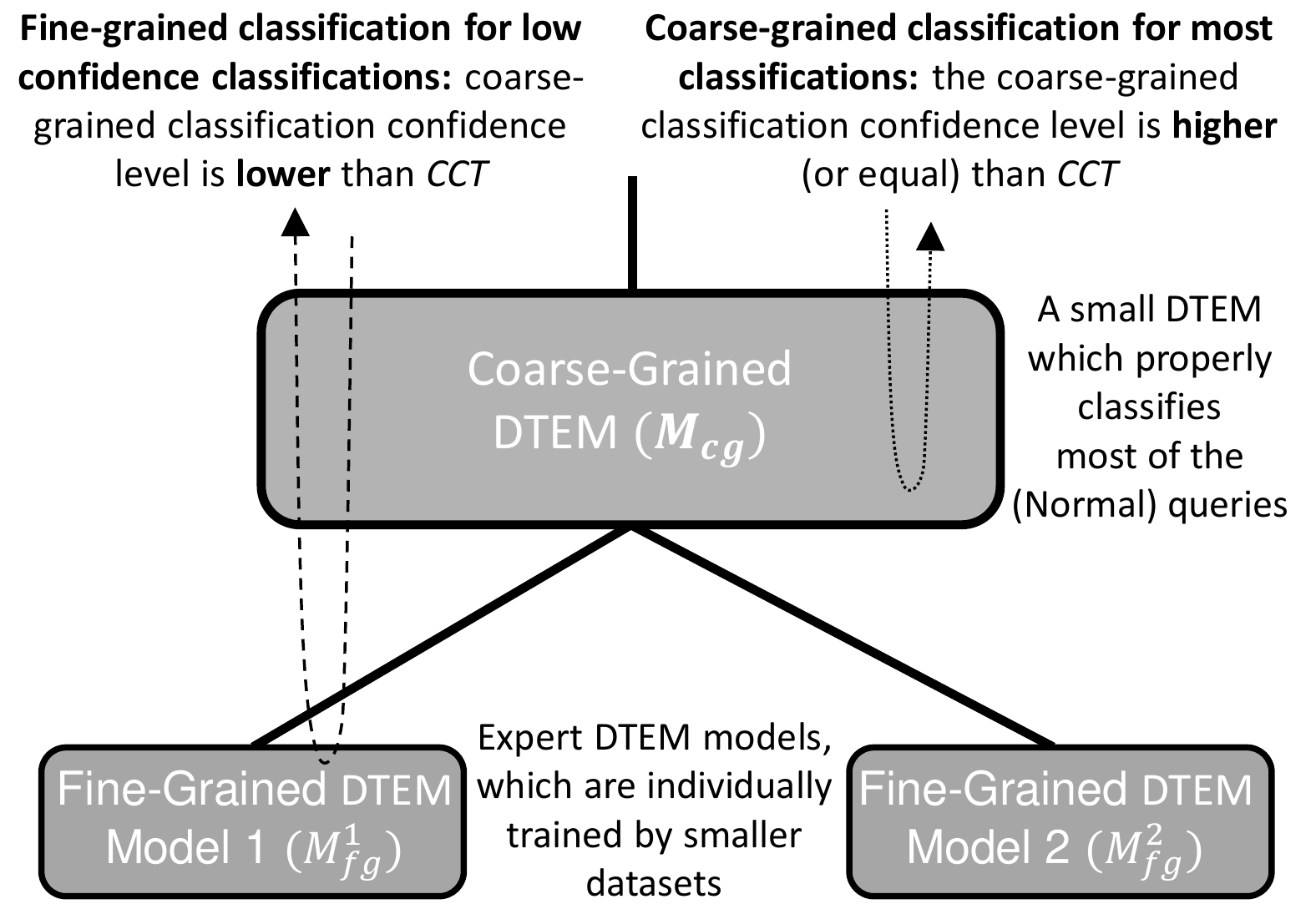}
		\caption{\name architecture.}
		\label{fig:model:a}
	\end{subfigure}
	
	\begin{subfigure}{0.95\linewidth}
		\includegraphics[width=\textwidth]{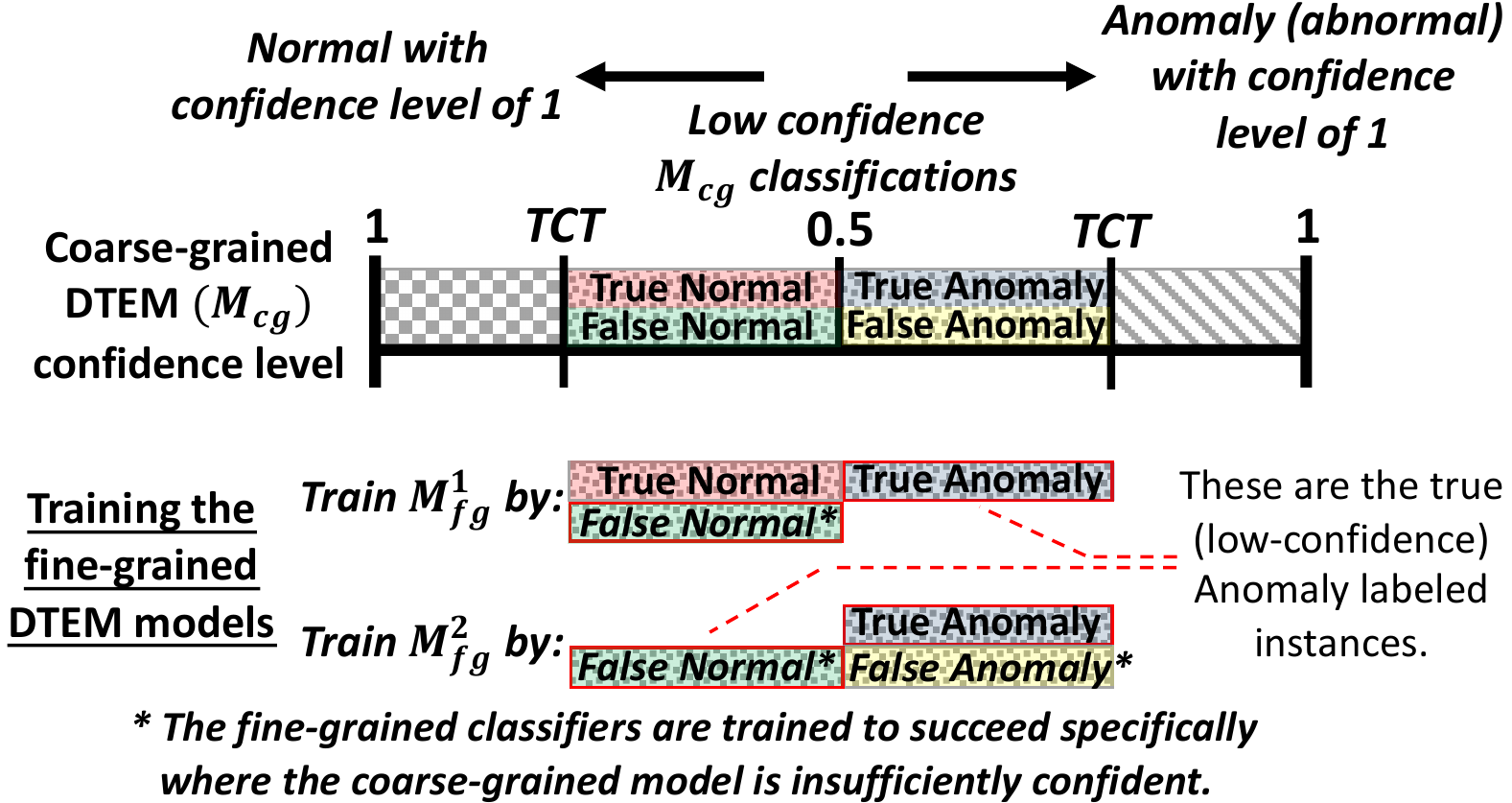}
		\caption{\name confidence level driven training.}
		\label{fig:model:b_trn}
		\vspace{6pt}
	\end{subfigure}

	\begin{subfigure}{0.88\linewidth}
		\includegraphics[width=\textwidth]{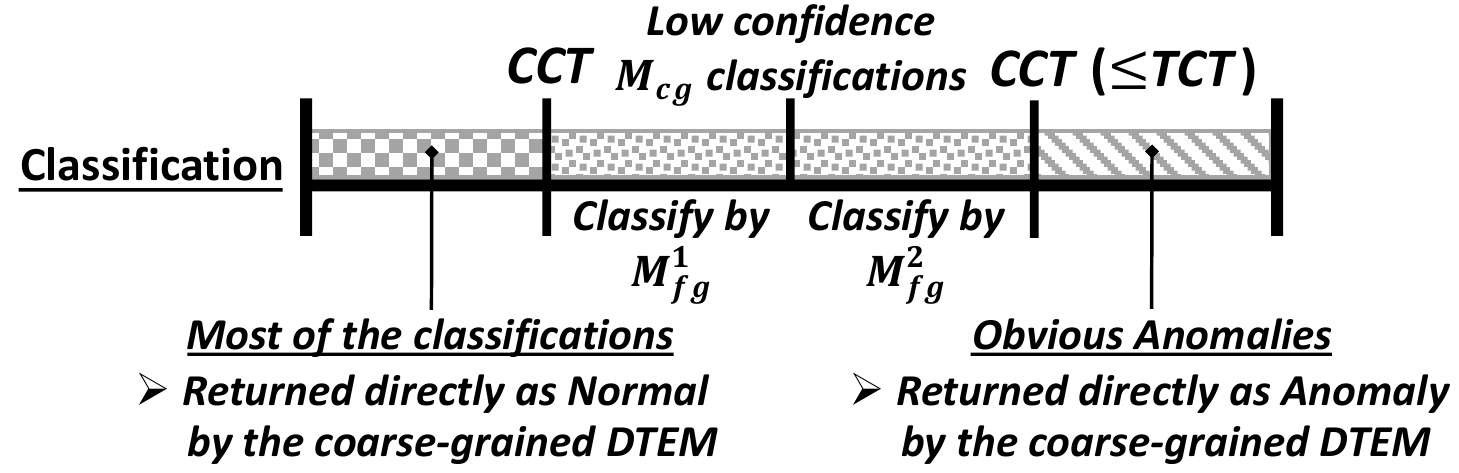}
		\caption{\name confidence level driven classification.}
		\label{fig:model:b_clf}
	\end{subfigure}

	\caption{\name architecture - a small (\cg) model and two expert (\fg) models. The training confidence threshold (\tct) at the coarse-grained model determines the training data for each \fg classifier. For classification, the classification confidence threshold (\cct) at the \cg model determines whether a classification query takes the short or the long path and which \fg model is queried for the latter case.}
	
	\vspace{-12pt}
	\label{fig:model}
	
\end{figure}


\subsection{The intuition behind \name{}}

So far, we have mainly discussed the ML-performance of \name{} and why, intuitively, it is expected to result in a high $F_1$ score. Now we discuss further intuition to why \name{} also results in lower memory footprint, lower training time (and hence lower energy consumption) and lower classification latency, as compared to the baseline models.

The training time complexity of DTEM methods depends on the size of the training dataset (\ie the number of training instances), the number of features, the number of trees and their depth limitations (if there are any). Whereas, the classification latency mostly depends on the model size (\ie the number of trees, and their depths).

Clearly, the smaller model size of a \cg classifier directly improves all of the criteria mentioned above. Additionally, as mentioned, the \fg classifiers are being trained by smaller datasets, which reduces their training time, and, often, their size. Indeed, our evaluation in Section \ref{sec:evaluation} shows that the size of the \fg models is smaller as compared to their corresponding baseline model for all tested data sets and classification methods.

Essentially, when considering a \name{} model (\ie both the \cg and \fg models), the classification latency of \name{} equals to a weighted average of the latencies according to the fraction of the classifications that are served by the \cg and \fg models. Since the \cg model serves most of the classifications, the averaged classification latency is expected to significantly improved as compared to the baseline model. Furthermore, our evaluation shows that the worst-case classification latency of \name{} (\ie the latency of a query that takes the longest path of \cg model and then the slowest \fg model) is also competitive and often lower than the latency of its corresponding baseline model
(for more details, see the evaluation in Section~\ref{sec:evaluation}).

On the other hand, the training time and model size of \name{} equal to the corresponding sums of the \cg and \fg models. Nevertheless, as presented in Section \ref{sec:evaluation}, our evaluation indicates that \name reduces the memory footprint and training time, as compared to the baseline models. 

To summarize, both the \cg and \fg models contribute to the overall improvements of \name{}, in the following ways: The \cg model is (1) based on a small classifier and, (2) serves most of the classification queries. The \fg models are (1) being trained by smaller data sets, (2) smaller as compared to the corresponding baseline and, (3) serve only a small fraction of the classification queries.

\section{\name}
\label{sec:system-label}

\T{Training.} Algorithm \ref{alg:train} describes the procedure for training a \name model. It begins with the training of a coarse-grained model (denoted by $M_{cg}$) using the entire labeled dataset. Next, we train the fine-grained models (denoted by $M_{fg}^i$ for \fg model $i$). To that end, we classify the labeled dataset by the coarse-grained model (line 5). Then, if the confidence level of the predicted top-1 class (\ie $max(d_x)$) is lower than the given training threshold (\tct), the labeled data instance is forwarded to both experts if its label is Anomaly (lines 7-9) or otherwise to a single fine-grained model according to the prediction made by $M_{cg}$ (lines 11-15). Notice that in lines 11-15, the data instances are forwarded \emph{according to their low-confidence coarse-grained classification, and not according to their labels}.
The reason is that we train the fine-grained models to succeed specifically where the coarse-grained model is insufficiently confident and is more likely to make a mistake. 

More specifically, as illustrated in Figure~\ref{fig:model:b_trn}, the data instances that are forwarded to fine-grained model 1 contain: (1) \textit{all} low confidence Anomaly instances and, (2) low confidence Normal instances that are correctly classified by the \cg model. Intuitively, this model becomes an expert in distinguishing between Normal instances that are correctly classified by the \cg model and Anomaly instances. Likewise, the data instances that are forwarded to fine-grained model 2 contain: (1) \textit{all} low confidence Anomaly instances and, (2) low confidence Normal instances that are misclassified by the \cg model. Intuitively, this model becomes an expert in distinguishing between misclassified Normal instances by the \cg model and Anomaly instances.  

The Anomaly class is significantly smaller (usually by orders of magnitude) than the Normal class in terms of the number of instances, and its low-confidence subset is even smaller. This fact results in two potential drawbacks, which we mitigate by the duplication of the low-confidence Anomaly labeled instances to both \fg models (lines 8-9), as explained in the following:

\noindent\textit{Less accurate Anomaly \cg classification:} Since the \cg model is trained using a rather small number of Anomaly instances as compared to the number of Normal instances, its classifications over these instances are very noisy with many misclassifications as compared to the Normal instances (see Figure \ref{fig:coarse_grained}). Namely, the classification distribution vector over these instances has a significant variance, which is even more severe for the low-confidence Anomaly subset. This makes the classification of the \cg model as to which \fg model we need to send a specific low-confidence Anomaly instance less reliable (unlike for the Normal instances).

\noindent\textit{Increased overfitting likelihood by the \fg models:} Due to the small cardinality of the low-confidence subset of the Anomaly instances, it is more likely for a \fg classifier to receive a non-sufficient number of such instances for training. This, in turn, increases the likelihood of overfitting the model. That is, it is more likely for the training of a \fg classifier to terminate in a state in which it has a nearly perfect $F_1$ score for the low-confidence subset of the Anomaly instances that were forwarded by the \cg model for its training, but this \fg model is likely to be less accurate at classification of a low-confidence Anomaly instance that may have been sent to the wrong \fg model and thus is more likely to be too different from other labeled instances this \fg classifier was trained on.

Therefore, by forwarding all the low-confidence subset of Anomalies to both experts, as we empirically find in our evaluations, reduces the likelihood of both drawbacks and makes the fine-grained models better experts for those queries in which the \cg model is more likely to make a classification mistake (\ie Observation 2). 

Note that this \textit{duplication} (\ie lines 8-9) results in a very low overhead in terms of the number of data instances used for the \fg models training (see Table~\ref{tab:main_results_table}).

\begin{algorithm}[t!]
\small
\caption{\name training}

\textbf{Input:} Labeled training data set $X$, confidence level \tct.\\
\textbf{ 1:\,\,} train $M_{cg}$ using $X$\\
\textbf{ 2:\,\,} set: $data[\mbox{fg}_1] = \emptyset$\\
\textbf{ 3:\,\,} set: $data[\mbox{fg}_2] = \emptyset$\\
\textbf{ 4:\,\,} for each $x \in X$:\\
\textbf{ 5:\,\,}\quad\quad\quad obtain coarse-grained distribution: $d_x$ = $M_{cg}(x)$\\ 
\textbf{ 6:\,\,}\quad\quad\quad if $max(d_x) < \tct$:\\
\textbf{ 7:\,\,}\quad\quad\quad\quad\quad\quad if $x.label == Anomaly$:\\
\textbf{ 8:\,\,}\quad\quad\quad\quad\quad\quad\quad\quad\quad update: $data[\mbox{fg}_1].append(x)$\\
\textbf{ 9:\,\,}\quad\quad\quad\quad\quad\quad\quad\quad\quad update: $data[\mbox{fg}_2].append(x)$\\
\textbf{10:}\quad\quad\quad\quad\quad\quad else:\\
\textbf{11:}\quad\quad\quad\quad\quad\quad\quad\quad\quad classify: $y$ = $argmax(d_x)$\\
\textbf{12:}\quad\quad\quad\quad\quad\quad\quad\quad\quad if $y == Normal$:\\
\textbf{13:}\quad\quad\quad\quad\quad\quad\quad\quad\quad\quad\quad\quad update: $data[\mbox{fg}_1].append(x)$\\
\textbf{14:}\quad\quad\quad\quad\quad\quad\quad\quad\quad else:\\
\textbf{15:}\quad\quad\quad\quad\quad\quad\quad\quad\quad\quad\quad\quad update: $data[\mbox{fg}_2].append(x)$\\
\textbf{16:} train $M_{fg}^{1}$ using $data[\mbox{fg}_1]$\\
\textbf{17:} train $M_{fg}^{2}$ using $data[\mbox{fg}_2]$\\
\label{alg:train}
\end{algorithm}

\T{Classification.} Algorithm \ref{alg:test} describes the procedure for a classification by \name, and Figure~\ref{fig:model:b_clf} illustrates it. First, we classify an arriving data instance by the coarse-grained model (lines 1-2). Whenever the resulting confidence level of the top-1 classification is greater than or equal to the classification confidence threshold, \cct, the classification by the coarse-grained model is \textit{valid} and therefore returned (line 4). As shown in Figure \ref{fig:coarse_grained}, we empirically find that most of the data instances result in a high confidence level. Therefore, since the size of the \cg model is small, these classification instances experience an extremely low-latency and high-throughput classification. The remaining small fraction of the data instances whose \cg classification is not valid (\ie which their resulting confidence level is lower than \cct) is forwarded to one of the fine-grained models, which is chosen according to the coarse-grained classification (lines 6-9). Specifically, if the \cg low-confidence classification is Normal, then the instance is forwarded to \fg model 1 which is trained to distinguish between Normal instances that are correctly classified by the \cg model and Anomaly instances (see Figure~\ref{fig:model:b_trn}). Likewise, if the \cg (low-confidence) classification is Anomaly, then the instance is forwarded to \fg model 2 which is trained to distinguish between Normal instances that are misclassified by the \cg model and Anomaly instances.

\begin{algorithm}[t!]
	\small
	\caption{\name classification}
	
	\textbf{Input:} Unlabeled data point $x$, confidence level \cct.\\
	\textbf{1:} obtain coarse-grained distribution: $d_x$ = $M_{cg}(x)$\\ 
	\textbf{2:} classify: $y$ = $argmax(d_x)$\\
	\textbf{3:} if $max(d_x) \ge \cct$: \\
	\textbf{4:}\quad\quad\quad return $y$\\
	\textbf{5:} else:\\
	\textbf{6:} \quad\quad\quad $y == Normal$ ? $c=1$ : $c=2$ \\
	\textbf{7:}\quad\quad\quad obtain fine-grained distribution: $\bar{d}_x$ = $M_{fg}^{c}$\\ 
	\textbf{8:}\quad\quad\quad classify: $\bar{y}$ = $argmax(\bar{d}_x)$\\
	\textbf{9:}\quad\quad\quad return $\bar{y}$\\
	\label{alg:test}
\end{algorithm}

\T{Putting it all together.} Figure \ref{fig:model} depicts a high-level architecture of \name, and the training and classification data-forwarding schemes. We may use \emph{different} confidence level thresholds for the fine-grained models training (\tct) and the classification/anomaly detection (\cct), such that ${\tct {\ge} \cct}$. The intuition for why it may be of interest to set $\tct{>}\cct$, is that it allows to train the fine-grained models with a bigger subset of the labeled data instances as compared to the classification subset that is forwarded to them. Tuning \tct{}, as we empirically find in our evaluations, often improves the anomaly detection capabilities (in terms of $F_1$ score) for a modest price in training time and \fg model sizes. 

Intuitively, the fine-grained models provide better classification and hence better anomaly detection for the uncertain classifications than the coarse-grained model for the following reasons: (1) we allow the fine-grained models to have more resources as compared to the coarse-grained model (see Section~\ref{subsec:tuning} for more details); (2) the fine-grained classifiers are trained by a much smaller fraction of the labeled training data. Essentially, each fine-grained model becomes an expert for its corresponding labeled data fraction, which represents the uncertain (and some of the wrong) classifications by the coarse-grained model. Note that when a classification query is forwarded to a \fg model, it is solely determined by this \fg model.

\T{\name{} model.} 
For ease of exposition, we define a specific configuration of a \name{} model by a tuple $R(C(T_{cg},D_{cg}), C(T_{fg},D_{fg}), \cct, \tct)$, that states the size limitation of the \cg{} model followed by the size limitation of the \fg{} models and finally the classification and training confidence thresholds.

\section{Implementation}

\name{} implementation is written in Python 3.6 and is based on the \textit{scikit-learn} library\footnote{v0.21.3 - released on July 30, 2019 (current stable version).}~\cite{scikit-learn}.
\name{} can augment any scikit-based DTEM classifier.
Specifically, we execute \name{} for internal scikit classifiers (RandomForest~\cite{scikit-rf}, GradientBoosting~\cite{scikit-gbdt}, and AdaBoost~\cite{scikit-adaboost}), as well as for Python packages (lightgbm-v2.2.3~\cite{python-lightgbm}, and xgboost-v0.90~\cite{python-xgboost}).

\section{Evaluation}
\label{sec:evaluation}

\subsection{Datasets}
\label{subsec:datasets}
As mentioned, we use three different datasets, where all are widely used, publicly available, and reproducible. To establish consistency for \name{}, we chose the dataset such that they are all from different areas and use-cases as well as with different levels of skewness (\ie the Normal to Anomaly number of instances ratio).

\T{KDD \cite{kdd_dataset}.} This dataset is a popular benchmark and is widely used for evaluation of IDS systems \cite{dhanabal2015study,kayacik2005selecting,sabhnani2003application}. It was used for The Third International Knowledge Discovery and Data Mining Tools Competition in which the task was to build a network intrusion detector. This database contains a standard set of data to be audited, which includes a wide variety of intrusions simulated in a military network environment. In our evaluation, we treat all intrusions (\eg DoS, Probe, R2L) as Anomalies ($24.389\%‬$) and all non-hostile connections as Normal. 

\T{Credit Card Fraud (CCF) \cite{ccf_dataset}.} This is a popular dataset that is used for anomaly and fraud detection benchmarking. \cite{ccf_req_cite_1,ccf_req_cite_4,ccf_req_cite_5,ccf_req_cite_7,ccf_req_cite_8}. The datasets contains transactions made by credit cards in September 2013 by european cardholders. This dataset presents transactions that occurred in two days, with frauds which we treat as Anomalies ($0.172\%$) where all the rest are legitimate transfers and treated as Normal class.

\T{Forest Cover (FC) \cite{forest_cover}.} This dataset is used in predicting forest cover type from cartographic variables \cite{blackard2000comparison,gama2003accurate,oza2001experimental,obradovic2004challenges}. This study includes four wilderness areas located in the Roosevelt National Forest of northern Colorado. 
Class 2 is considered as Normal, and class 4 as Anomaly ($0.9\%$).

\subsection{Tuning \name{}}
\label{subsec:tuning}

The tuning of a \name{} classifier starts by identifying a set of sensible candidate hyperparameters for the \cg model. Recall that we want a small \cg model that can make valid classifications for most of the data instances with a high $F_1$ score (\eg Figure \ref{fig:coarse_grained}). 

To that end, a set of such sensible hyperparameters may be derived by looking at known default configurations and thumb rules for a specific standard full-sized classifier and for similar datasets (if such are known) and considering smaller sized options such that the resulting \cg model will be smaller than a standard full-sized classifier by some constant factor (\eg 5$\times$).

The candidate hyperparameters for the \fg models are then chosen similarly but by considering larger sizes (\ie number of trees, depth limitations) varying from the \cg and up to a standard model size.

Next, we need to consider candidate \cct{} and \tct{} values. The set of interest is always $0.5\le\cct{}\le\tct{}\le1$ sampled according to some granularity (\eg 0.05).

Finally, once we have our grid of sensible hyperparameter configurations for \name{}, we perform iterations of our 5-fold cross-validation process, each time using different model settings from this predetermined set. This grid-search tuning process is no different from standard practice. 

Clearly, these parameters are not guaranteed to be optimal. With that said, finding satisfactory hyperparameters often requires an extensive grid-search, like any other classifier.

\begin{figure}[t!]
	\centering
	
	\begin{subfigure}{\linewidth}
		\includegraphics[width=\textwidth,trim={1.0cm 7cm 0 0},clip]{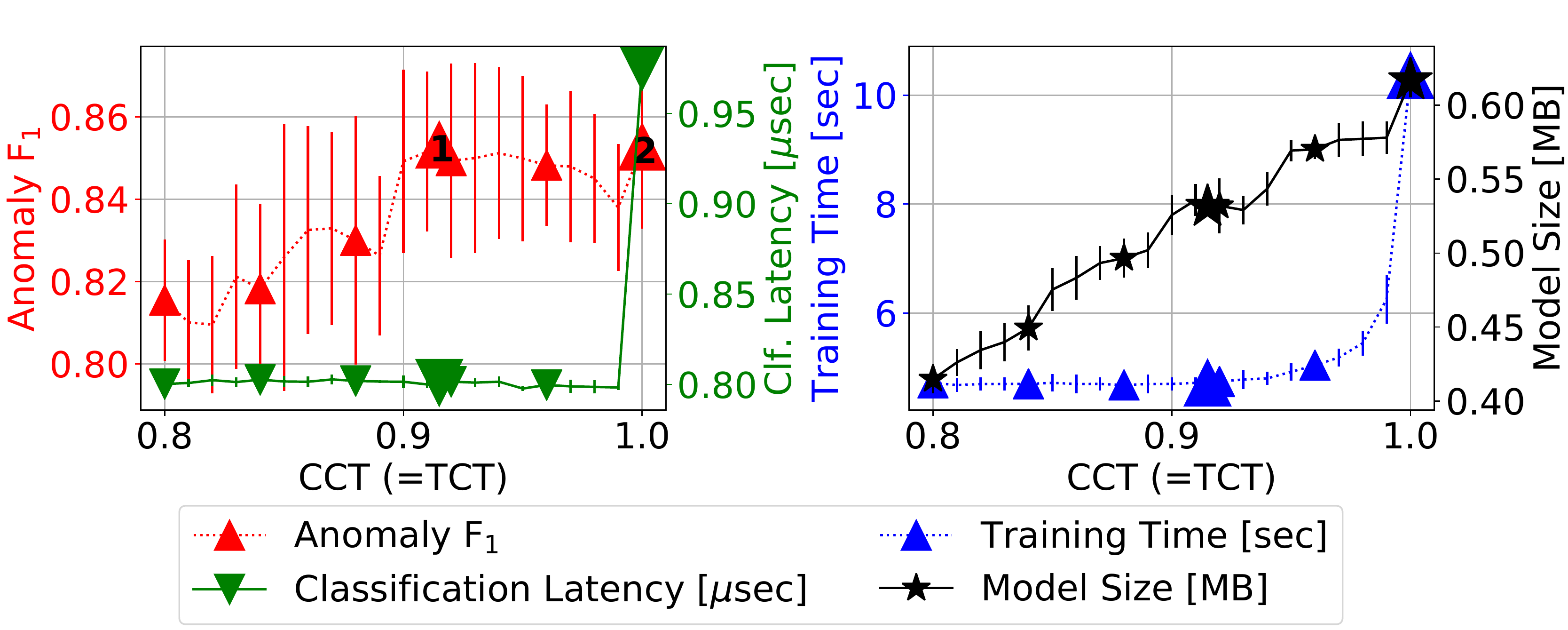}
		\caption{How \cct{} and \tct{}(=\cct{}) affect different metrics. The two \cct{} values that achieve the highest Anomaly $F_1$ are marked.}
		\label{fig:cct_sweep_figure}
		\vspace{5pt}
	\end{subfigure}
	\begin{subfigure}{\linewidth}
	
        \resizebox{\linewidth}{!}{%
        \begin{tabular}{|c|c|c|c|c|c|c|}
        \hline
        \textbf{Index} & \textbf{CCT} & \textbf{\begin{tabular}[c]{@{}c@{}}Anom-\\aly $F_1$ \end{tabular}} & \textbf{\begin{tabular}[c]{@{}c@{}}Fine\\Grained\\(FG) train\\ data\\ fraction\end{tabular}} & \textbf{\begin{tabular}[c]{@{}c@{}}FG\\ Nor./Anom. \\ train data\\ ratio \\ {[}$M_{fg}^1$, $M_{fg}^2${]}\end{tabular}} & \textbf{\begin{tabular}[c]{@{}c@{}}FG \\ total\\ test data\\ fraction\end{tabular}} \\ \hline
        \textbf{-} & 0.5   & 0.810   & 0.0\%  & nan, nan & 0.0\% \\ \hline
        \textbf{1} & 0.915 & 0.853   & 2.32\% & 45.11, 0.348 & 2.34\% \\ \hline
        \textbf{2} & 1.0   & 0.852   & 98.4\% & 567.7, 0.136 & 98.2\% \\ \hline
        \end{tabular}%
        }
		\caption{\name attributes for the marked \cct{} values in (a), and for \cct{}=0.5 -- \ie when \textit{all} \cg model classifications are considered as valid. Note that the original Normal to Anomaly instance ratio of CCF is 581.4, and the Anomaly $F_1$ of the corresponding baseline model is 0.845471.}
		\label{tab:cct_sweep}        
	\end{subfigure}

	\caption{How \cct{} and \tct{}(=\cct{}) affect \name, demonstrated with RF-based $R(C(20,10), C(25,10), -, -)$ for CCF dataset.}
	
	\vspace{-14pt}
	\label{fig:sweep_example}
	
\end{figure}

\textit{Example.} 
We next demonstrate how \cct{} and \tct{} affect the $F_1$ score, classification latency, training time, and model size. 
Figure~\ref{fig:cct_sweep_figure} presents these metrics vs. different values of \cct{} (with \tct{}=\cct{} for simplicity) of a RF-based $R(C(20,10), C(25,10), -, -)$ for the CCF dataset. 
We identify two \cct{} values, \cct{}=0.915 and \cct{}=1, that achieve the highest Anomaly $F_1$ (marked in the graph). 
Notice that, \cct{}=1 results in forwarding \textit{most} of the test dataset to the \fg models, which in turn increases the classification latency. 
Moreover, as \tct{} increases, a higher fraction of the training-data is being forwarded to the \fg models, which in turn increases both the model size and training-time.

Table~\ref{tab:cct_sweep} also presents the Normal to Anomaly ratio of the training data instances that are forwarded to each of the \fg models. As can be seen, this ratio is further improved as compared to the original dataset ratio (581.4) for \fg model 1 with \cct{}=0.915 (while \cct{}=1 results in a ratio similar to the original one). Notice that, for \fg model 2, the ratio is lower than one, which means that the model is being trained with more Anomaly-labeled instances than Normal-labeled ones. Additionally, the table presents these attributes for $R(C(20,10), C(25,10), 0.5, 0.5)$ model, which is essentially identical to $C(20,10)$ (a smaller baseline model), that returns \textit{all} of the data instance classifications \textit{only} by the \cg model. This \name{} model demonstrates that the \fg models are essential to achieve sufficient Anomaly $F_1$, even when their total test dataset fraction is relatively low (\eg $2.34\%$ for \cct{}=0.915).

A better \cct{} and \tct{} configuration for this \name{} model, $R(C(20,10), C(25,10), 0.995, 1.0)$, is achieved by the grid-search process mentioned above, as further detailed next in Table~\ref{tab:main_results_table}.

\begin{table*}[]
\centering
\vspace{-6pt}
\caption{Comparison among two \name{} configurations to the baseline over three classification DTEM methods, each with three different datasets. All results are obtained using 5-fold cross-validation. \name{} achieves competitive and often superior $F_1$ score with lower training time, classification latency, and superior (for bagging) or competitive (for boosting) model size.}
\vspace{-6pt}
\label{tab:main_results_table}
\resizebox{\textwidth}{!}{%
\begin{tabular}{|c|c|c|l|c|c|c|c|c|c|c|c|}
\hline
Classifier & Dataset & \# & \multicolumn{1}{c|}{Model} & \begin{tabular}[c]{@{}c@{}}Normal\\ $F_1$\end{tabular} & \begin{tabular}[c]{@{}c@{}}Anomaly\\ $F_1$\end{tabular} & \begin{tabular}[c]{@{}c@{}}Model\\ size\\ {[}MB{]}\end{tabular} & \begin{tabular}[c]{@{}c@{}}Train\\ time\\ {[}s{]}\end{tabular} & \begin{tabular}[c]{@{}c@{}}Fine-grained\\ train data\\ \% {[}fg1, fg2{]}\end{tabular} & \begin{tabular}[c]{@{}c@{}}Classification\\ latency\\ {[}$\mu s${]}\end{tabular} & \begin{tabular}[c]{@{}c@{}}RADE\\ worst-case\\ classification\\ latency {[}$\mu s${]}\end{tabular} & \begin{tabular}[c]{@{}c@{}}Fine-grained\\ test data\\ \% {[}fg1, fg2{]}\end{tabular} \\ \hline
 &  & \cellcolor[HTML]{C0C0C0}1 & \cellcolor[HTML]{C0C0C0}Baseline -- $C(150,None)$ & \cellcolor[HTML]{C0C0C0}0.999913 & \cellcolor[HTML]{C0C0C0}0.995190 & \cellcolor[HTML]{C0C0C0}5.57 & \cellcolor[HTML]{C0C0C0}22.5 & \cellcolor[HTML]{C0C0C0}-- & \cellcolor[HTML]{C0C0C0}9.4 & \cellcolor[HTML]{C0C0C0}-- & \cellcolor[HTML]{C0C0C0}-- \\ \cline{3-12} 
 &  & 2 & $R(C(10,10),C(20,20),0.98,0.995)$ & 0.999910 & 0.995028 & 1.02 & 3.5 & 9.55\%, 0.57\% & 1.3 & 6.9 & 5.19\%, 0.37\% \\ \cline{3-12} 
 & \multirow{-3}{*}{KDD} & 3 & $R(C(10,10),C(35,20),0.98,0.98)$ & 0.999911 & 0.995091 & 1.54 & 3.6 & 5.38\%, 0.35\% & 1.3 & 10.1 & 5.19\%, 0.37\% \\ \cline{2-12} 
 &  & \cellcolor[HTML]{C0C0C0}4 & \cellcolor[HTML]{C0C0C0}Baseline -- $C(85,None)$ & \cellcolor[HTML]{C0C0C0}0.999759 & \cellcolor[HTML]{C0C0C0}0.845471 & \cellcolor[HTML]{C0C0C0}2.01 & \cellcolor[HTML]{C0C0C0}75.6 & \cellcolor[HTML]{C0C0C0}-- & \cellcolor[HTML]{C0C0C0}6.2 & \cellcolor[HTML]{C0C0C0}-- & \cellcolor[HTML]{C0C0C0}-- \\ \cline{3-12} 
 &  & 5 & $R(C(20,10),C(30,20),0.94,0.953)$ & 0.999733 & 0.840198 & 0.82 & 20.3 & 6.82\%, 0.09\% & 1.9 & 4.7 & 3.72\%, 0.06\% \\ \cline{3-12} 
 & \multirow{-3}{*}{CCF} & 6 & $R(C(20,10),C(25,10),0.995,1.0)$ & 0.999756 & 0.854553 & 0.63 & 40.7 & 97.57\%, 0.20\% & 2.1 & 3.4 & 41.78\%, 0.09\% \\ \cline{2-12} 
 &  & \cellcolor[HTML]{C0C0C0}7 & \cellcolor[HTML]{C0C0C0}Baseline -- $C(80,None)$ & \cellcolor[HTML]{C0C0C0}0.999379 & \cellcolor[HTML]{C0C0C0}0.859186 & \cellcolor[HTML]{C0C0C0}9.95 & \cellcolor[HTML]{C0C0C0}16.7 & \cellcolor[HTML]{C0C0C0}-- & \cellcolor[HTML]{C0C0C0}5.2 & \cellcolor[HTML]{C0C0C0}-- & \cellcolor[HTML]{C0C0C0}-- \\ \cline{3-12} 
 &  & 8 & $R(C(20,15),C(35,20),1,1.0)$ & 0.999462 & 0.883542 & 5.96 & 6.9 & 6.28\%, 1.23\% & 1.9 & 8.1 & 5.81\%, 1.24\% \\ \cline{3-12} 
\multirow{-9}{*}{RF} & \multirow{-3}{*}{FC} & 9 & $R(C(20,20),C(25,None),0.99,0.9966)$ & 0.999438 & 0.882269 & 4.66 & 6.3 & 2.76\%, 0.58\% & 1.8 & 7.9 & 2.46\%, 0.58\% \\ \hline
 &  & \cellcolor[HTML]{C0C0C0}10 & \cellcolor[HTML]{C0C0C0}Baseline -- $C(450,3)$ & \cellcolor[HTML]{C0C0C0}0.999933 & \cellcolor[HTML]{C0C0C0}0.996294 & \cellcolor[HTML]{C0C0C0}0.25 & \cellcolor[HTML]{C0C0C0}196.5 & \cellcolor[HTML]{C0C0C0}-- & \cellcolor[HTML]{C0C0C0}13.6 & \cellcolor[HTML]{C0C0C0}-- & \cellcolor[HTML]{C0C0C0}-- \\ \cline{3-12} 
 &  & 11 & $R(C(70,5),C(250,3),0.99,0.99444)$ & 0.999934 & 0.996356 & 0.32 & 49.4 & 0.33\%, 0.13\% & 3.6 & 11.4 & 0.12\%, 0.10\% \\ \cline{3-12} 
 & \multirow{-3}{*}{KDD} & 12 & $R(C(90,5),C(350,3),0.99,0.9988)$ & 0.999934 & 0.996357 & 0.40 & 65.3 & 0.70\%, 0.24\% & 4.7 & 15.1 & 0.08\%, 0.08\% \\ \cline{2-12} 
 &  & \cellcolor[HTML]{C0C0C0}13 & \cellcolor[HTML]{C0C0C0}Baseline -- $C(500,3)$ & \cellcolor[HTML]{C0C0C0}0.999803 & \cellcolor[HTML]{C0C0C0}0.877078 & \cellcolor[HTML]{C0C0C0}0.28 & \cellcolor[HTML]{C0C0C0}252.4 & \cellcolor[HTML]{C0C0C0}-- & \cellcolor[HTML]{C0C0C0}16.4 & \cellcolor[HTML]{C0C0C0}-- & \cellcolor[HTML]{C0C0C0}-- \\ \cline{3-12} 
 &  & 14 & $R(C(70,5),C(200,3),0.99,0.9977)$ & 0.999792 & 0.873624 & 0.26 & 61.7 & 0.82\%, 0.15\% & 3.9 & 13.0 & 0.15\%, 0.07\% \\ \cline{3-12} 
 & \multirow{-3}{*}{CCF} & 15 & $R(C(90,5),C(300,3),0.908,0.98)$ & 0.999800 & 0.878426 & 0.32 & 77.9 & 0.09\%, 0.05\% & 4.9 & 14.7 & 0.02\%, 0.02\% \\ \cline{2-12} 
 &  & \cellcolor[HTML]{C0C0C0}16 & \cellcolor[HTML]{C0C0C0}Baseline -- $C(2000,3)$ & \cellcolor[HTML]{C0C0C0}0.999495 & \cellcolor[HTML]{C0C0C0}0.891222 & \cellcolor[HTML]{C0C0C0}1.21 & \cellcolor[HTML]{C0C0C0}1367.9 & \cellcolor[HTML]{C0C0C0}-- & \cellcolor[HTML]{C0C0C0}58.9 & \cellcolor[HTML]{C0C0C0}-- & \cellcolor[HTML]{C0C0C0}-- \\ \cline{3-12} 
 &  & 17 & $R(C(70,5),C(300,5),0.99,0.995)$ & 0.999489 & 0.892526 & 0.85 & 79.4 & 1.94\%, 0.50\% & 3.9 & 20.5 & 1.13\%, 0.41\% \\ \cline{3-12} 
\multirow{-9}{*}{XGBoost} & \multirow{-3}{*}{FC} & 18 & $R(C(70,5),C(500,5),0.99,0.99)$ & 0.999506 & 0.896179 & 1.23 & 83.0 & 1.49\%, 0.48\% & 4.0 & 29.9 & 1.13\%, 0.41\% \\ \hline
 &  & \cellcolor[HTML]{C0C0C0}19 & \cellcolor[HTML]{C0C0C0}Baseline -- $C(700,3)$ & \cellcolor[HTML]{C0C0C0}0.999943 & \cellcolor[HTML]{C0C0C0}0.996854 & \cellcolor[HTML]{C0C0C0}0.92 & \cellcolor[HTML]{C0C0C0}874.6 & \cellcolor[HTML]{C0C0C0}-- & \cellcolor[HTML]{C0C0C0}136.6 & \cellcolor[HTML]{C0C0C0}-- & \cellcolor[HTML]{C0C0C0}-- \\ \cline{3-12} 
 &  & 20 & $R(C(300,3),C(500,5),0.64,0.84)$ & 0.999955 & 0.997485 & 2.48 & 630.6 & 99.96\%, 1.76\% & 61.9 & 165.6 & 3.67\%, 0.47\% \\ \cline{3-12} 
 & \multirow{-3}{*}{KDD} & 21 & $R(C(300,3),C(450,6),0.669,0.742)$ & 0.999947 & 0.997084 & 2.69 & 462.4 & 59.23\%, 1.51\% & 64.8 & 149.3 & 12.53\%, 0.86\% \\ \cline{2-12} 
 &  & \cellcolor[HTML]{C0C0C0}22 & \cellcolor[HTML]{C0C0C0}Baseline -- $C(1100,2)$ & \cellcolor[HTML]{C0C0C0}0.999801 & \cellcolor[HTML]{C0C0C0}0.876765 & \cellcolor[HTML]{C0C0C0}0.99 & \cellcolor[HTML]{C0C0C0}2192.4 & \cellcolor[HTML]{C0C0C0}-- & \cellcolor[HTML]{C0C0C0}185.9 & \cellcolor[HTML]{C0C0C0}-- & \cellcolor[HTML]{C0C0C0}-- \\ \cline{3-12} 
 &  & 23 & $R(C(200,3),C(450,6),0.649,0.649)$ & 0.999798 & 0.877588 & 2.41 & 716.0 & 7.60\%, 0.12\% & 39.6 & 118.4 & 7.52\%, 0.10\% \\ \cline{3-12} 
 & \multirow{-3}{*}{CCF} & 24 & $R(C(450,3),C(350,6),0.673,0.855)$ & 0.999808 & 0.882403 & 2.31 & 3048.6 & 96.70\%, 0.17\% & 78.7 & 135.6 & 14.38\%, 0.11\% \\ \cline{2-12} 
 &  & \cellcolor[HTML]{C0C0C0}25 & \cellcolor[HTML]{C0C0C0}Baseline -- $C(1000,3)$ & \cellcolor[HTML]{C0C0C0}0.999449 & \cellcolor[HTML]{C0C0C0}0.882856 & \cellcolor[HTML]{C0C0C0}1.44 & \cellcolor[HTML]{C0C0C0}1333.2 & \cellcolor[HTML]{C0C0C0}-- & \cellcolor[HTML]{C0C0C0}235.8 & \cellcolor[HTML]{C0C0C0}-- & \cellcolor[HTML]{C0C0C0}-- \\ \cline{3-12} 
 &  & 26 & $R(C(300,4),C(300,6),0.675,0.855)$ & 0.999477 & 0.892222 & 2.17 & 877.5 & 57.29\%, 0.47\% & 73.2 & 147.6 & 4.33\%, 0.46\% \\ \cline{3-12} 
\multirow{-9}{*}{AdaBoost} & \multirow{-3}{*}{FC} & 27 & $R(C(60,3),C(350,5),0.68,0.715)$ & 0.999472 & 0.889118 & 3.10 & 115.9 & 5.99\%, 0.79\% & 17.8 & 104.1 & 3.87\%, 0.79\% \\ \hline
\end{tabular}%
}
\vspace{-12pt}
\end{table*}
\subsection{\name vs. standard methods}
\label{subsec:evaluationcompare}
We compare \name{} to the baseline models over an AWS m5d.16xlarge EC2 instance with Ubuntu 16.04 OS~\cite{awsec2}, and summarize the evaluation results in Table \ref{tab:main_results_table}. 
The results of RF, XGBoost and AdaBoost are summarized, while similar improvements are obtained for GBDT and LightGBM.
For each classifier and dataset, we present results for the baseline model and two different configurations of \name{}. We rely on 5-fold cross-validation and report the mean values. 

\subsubsection{Bagging - Random Forest} 

\T{Anomaly detection.} For all three datasets \name{} exhibits competitive or superior $F_1$ scores as compared to the baseline\footnote{As mentioned, we consider $F_1$ score as the ML performance measure for anomaly detection (\ie imbalanced datasets). Nevertheless, we also evaluate \name{} models for AUC and Average Precision and reach similar conclusions.}. Specifically, the results are somewhat similar for the KDD and CCF; whereas for FC, both \name{} configurations result in an advantage of $\approx +2.4\%$ in Anomaly $F_1$.

\T{Model size.} All \name{} model sizes are notably smaller than their corresponding baseline model. For example, for KDD, $R(C(10,10),C(20,20),0.98,0.995)$ is $5.46\times$ smaller than $C(150,None)$.

\T{Training time.} For all three datasets, the training time of \name{} is significantly lower. For example, it is $6.42\times$ faster for KDD. These lower training times come in line with the smaller size of the \cg model as compared to the baseline and the small fractions of the training data that are used for the training of the \fg models.

\T{Classification latency.} The improvement of \name{} over the baseline is consistent and is up to $7.23\times$ faster due to the smaller latency introduced by the \cg model. Even when considering the worst-case classification latency for \name{} (\ie a query that takes the path of \cg model and then the slowest \fg model) \name{} is still competitive. The non-negligible difference between the average and worst-case classification latency for \name{} falls in line with the small fractions of queries taking the long path (\eg 5.56\% for both \fg models for KDD). 

\begin{table}[]
\centering
\caption{Raspberry Pi 3 - Training-time and classification latency comparison between \name{} and the baseline models.
}
\vspace{-4pt}
\resizebox{\linewidth}{!}{%
\begin{tabular}{cc|c|c|c|c|c|c|c|c|}
\cline{3-4} \cline{6-7} \cline{9-10}
 &  & \multicolumn{2}{c|}{RF} &  & \multicolumn{2}{c|}{XGBoost} &  & \multicolumn{2}{c|}{AdaBoost} \\ \cline{2-10} 
\multicolumn{1}{c|}{} & \cellcolor[HTML]{EFEFEF}\# & \begin{tabular}[c]{@{}c@{}}Training\\ time {[}s{]}\end{tabular} & \begin{tabular}[c]{@{}c@{}}Classification\\ Latency {[}$\mu s${]}\end{tabular} & \cellcolor[HTML]{EFEFEF}\# & \begin{tabular}[c]{@{}c@{}}Training\\ time {[}s{]}\end{tabular} & \begin{tabular}[c]{@{}c@{}}Classification\\ Latency {[}$\mu s${]}\end{tabular} & \cellcolor[HTML]{EFEFEF}\# & \begin{tabular}[c]{@{}c@{}}Training\\ time {[}s{]}\end{tabular} & \begin{tabular}[c]{@{}c@{}}Classification\\ Latency {[}$\mu s${]}\end{tabular} \\ \hline
\multicolumn{1}{|c|}{} & \cellcolor[HTML]{EFEFEF}1 & \cellcolor[HTML]{C0C0C0}567 & \cellcolor[HTML]{C0C0C0}127 & \cellcolor[HTML]{EFEFEF}10 & \cellcolor[HTML]{C0C0C0}2068 & \cellcolor[HTML]{C0C0C0}166 & \cellcolor[HTML]{EFEFEF}19 & \cellcolor[HTML]{C0C0C0}11727 & \cellcolor[HTML]{C0C0C0}1384 \\ \cline{2-10} 
\multicolumn{1}{|c|}{} & \cellcolor[HTML]{EFEFEF}2 & \begin{tabular}[c]{@{}c@{}}63\\ (9.0$\times$)\end{tabular} & \begin{tabular}[c]{@{}c@{}}26\\ (4.9$\times$)\end{tabular} & \cellcolor[HTML]{EFEFEF}11 & \begin{tabular}[c]{@{}c@{}}608\\ (3.4$\times$)\end{tabular} & \begin{tabular}[c]{@{}c@{}}53\\ (3.1$\times$)\end{tabular} & \cellcolor[HTML]{EFEFEF}20 & \begin{tabular}[c]{@{}c@{}}12433\\ (0.9$\times$)\end{tabular} & \begin{tabular}[c]{@{}c@{}}619\\ (2.2$\times$)\end{tabular} \\ \cline{2-10} 
\multicolumn{1}{|c|}{\multirow{-3}{*}{\begin{tabular}[c]{@{}c@{}}K\\ D\\ D\end{tabular}}} & \cellcolor[HTML]{EFEFEF}3 & \begin{tabular}[c]{@{}c@{}}57\\ (10.0$\times$)\end{tabular} & \begin{tabular}[c]{@{}c@{}}22\\ (5.8$\times$)\end{tabular} & \cellcolor[HTML]{EFEFEF}12 & \begin{tabular}[c]{@{}c@{}}536\\ (3.9$\times$)\end{tabular} & \begin{tabular}[c]{@{}c@{}}47\\ (3.6$\times$)\end{tabular} & \cellcolor[HTML]{EFEFEF}21 & \begin{tabular}[c]{@{}c@{}}8021\\ (1.5$\times$)\end{tabular} & \begin{tabular}[c]{@{}c@{}}697\\ (2.0$\times$)\end{tabular} \\ \hline
\multicolumn{1}{|c|}{} & \cellcolor[HTML]{EFEFEF}4 & \cellcolor[HTML]{C0C0C0}848 & \cellcolor[HTML]{C0C0C0}89 & \cellcolor[HTML]{EFEFEF}13 & \cellcolor[HTML]{C0C0C0}2909 & \cellcolor[HTML]{C0C0C0}164 & \cellcolor[HTML]{EFEFEF}22 & \cellcolor[HTML]{C0C0C0}17524 & \cellcolor[HTML]{C0C0C0}1889 \\ \cline{2-10} 
\multicolumn{1}{|c|}{} & \cellcolor[HTML]{EFEFEF}5 & \begin{tabular}[c]{@{}c@{}}195\\ (4.4$\times$)\end{tabular} & \begin{tabular}[c]{@{}c@{}}41\\ (2.2$\times$)\end{tabular} & \cellcolor[HTML]{EFEFEF}14 & \begin{tabular}[c]{@{}c@{}}713\\ (4.1$\times$)\end{tabular} & \begin{tabular}[c]{@{}c@{}}49\\ (3.3$\times$)\end{tabular} & \cellcolor[HTML]{EFEFEF}23 & \begin{tabular}[c]{@{}c@{}}24905\\ (0.7$\times$)\end{tabular} & \begin{tabular}[c]{@{}c@{}}755\\ (2.5$\times$)\end{tabular} \\ \cline{2-10} 
\multicolumn{1}{|c|}{\multirow{-3}{*}{\begin{tabular}[c]{@{}c@{}}C\\ C\\ F\end{tabular}}} & \cellcolor[HTML]{EFEFEF}6 & \begin{tabular}[c]{@{}c@{}}433\\ (2.0$\times$)\end{tabular} & \begin{tabular}[c]{@{}c@{}}41\\ (2.2$\times$)\end{tabular} & \cellcolor[HTML]{EFEFEF}15 & \begin{tabular}[c]{@{}c@{}}892\\ (3.3$\times$)\end{tabular} & \begin{tabular}[c]{@{}c@{}}45\\ (3.7$\times$)\end{tabular} & \cellcolor[HTML]{EFEFEF}24 & \begin{tabular}[c]{@{}c@{}}5681\\ (3.1$\times$)\end{tabular} & \begin{tabular}[c]{@{}c@{}}425\\ (4.4$\times$)\end{tabular} \\ \hline
\multicolumn{1}{|c|}{} & \cellcolor[HTML]{EFEFEF}7 & \cellcolor[HTML]{C0C0C0}363 & \cellcolor[HTML]{C0C0C0}66 & \cellcolor[HTML]{EFEFEF}16 & \cellcolor[HTML]{C0C0C0}9519 & \cellcolor[HTML]{C0C0C0}1831 & \cellcolor[HTML]{EFEFEF}25 & \cellcolor[HTML]{C0C0C0}24355 & \cellcolor[HTML]{C0C0C0}2596 \\ \cline{2-10} 
\multicolumn{1}{|c|}{} & \cellcolor[HTML]{EFEFEF}8 & \begin{tabular}[c]{@{}c@{}}113\\ (3.2$\times$)\end{tabular} & \begin{tabular}[c]{@{}c@{}}28\\ (2.3$\times$)\end{tabular} & \cellcolor[HTML]{EFEFEF}17 & \begin{tabular}[c]{@{}c@{}}601\\ (15.8$\times$)\end{tabular} & \begin{tabular}[c]{@{}c@{}}68\\ (26.9$\times$)\end{tabular} & \cellcolor[HTML]{EFEFEF}26 & \begin{tabular}[c]{@{}c@{}}22971\\ (1.1$\times$)\end{tabular} & \begin{tabular}[c]{@{}c@{}}762\\ (3.4$\times$)\end{tabular} \\ \cline{2-10} 
\multicolumn{1}{|c|}{\multirow{-3}{*}{\begin{tabular}[c]{@{}c@{}}F\\ C\end{tabular}}} & \cellcolor[HTML]{EFEFEF}9 & \begin{tabular}[c]{@{}c@{}}116\\ (3.1$\times$)\end{tabular} & \begin{tabular}[c]{@{}c@{}}28\\ (2.3$\times$)\end{tabular} & \cellcolor[HTML]{EFEFEF}18 & \begin{tabular}[c]{@{}c@{}}588\\ (16.2$\times$)\end{tabular} & \begin{tabular}[c]{@{}c@{}}59\\ (30.8$\times$)\end{tabular} & \cellcolor[HTML]{EFEFEF}27 & \begin{tabular}[c]{@{}c@{}}2368\\ (10.3$\times$)\end{tabular} & \begin{tabular}[c]{@{}c@{}}203\\ (12.8$\times$)\end{tabular} \\ \hline
\end{tabular}%
}
\vspace{-15pt}
\label{tab:pi_table}
\end{table}

\subsubsection{Boosting - XGBoost and AdaBoost}
\label{subsubsec:boosting}
\T{Anomaly detection.} For all three datasets, \name{} exhibits competitive or superior anomaly detection capabilities as compared to the baseline. For example, for XGBoost over FC, \name{} configuration achieves a better Anomaly $F_1$ by $+0.5\%$; and
for AdaBoost over FC, \name{} configurations improve the Anomaly $F_1$ by $\approx +1\%$ and $ +0.6\%$. 

\T{Model size.} XGBoost model sizes of \name{} and the baseline are competitive. AdaBoost model sizes of \name{} are up to $2.92\times$ larger than the baseline. This is of less concern since AdaBoost DTEM is not memory bound (uses trees with only a few (\ie stumps) to several terminal nodes).

\T{Training time.} For all datasets, the training time of \name{} is significantly lower. For example, it is $17.2\times$ faster for XGBoost over FC. These lower times adhere with the smaller size of the \cg models the smaller  data fractions that are used for the training of the \fg models.

\T{Classification latency.} The improvement in the mean classification latency of \name{} over the baseline is even more significant than in the case of RF. For example, it is $15.1\times$ faster for XGBoost over FC and $13.24\times$ faster for AdaBoost over FC. The worst-case classification time of \name{} is also competitive as compared to the baseline and is even superior (by up to $2.83\times$ faster) for the CCF and FC datasets. Note that the classification latency grows linearly with respect to the number of trees but only logarithmic in the (roughly balanced) tree size (\ie number of terminal nodes). Therefore, the lower number of trees used by the \cg and \fg models of \name{} results in faster classification as compared to the baseline.  

\subsection{Raspberry Pi 3 evaluation}
We evaluate \name{} over Raspberry Pi 3 B+\footnote{ARM Cortex-A53 with 512 KB shared L2 and 1GB SDRAM.} with Raspbian 9.8 OS \cite{raspi}.
Table~\ref{tab:pi_table} compares the training time and classification latency of \name{} and baseline models, for the same classification DTEM methods and datasets, as in Table~\ref{tab:main_results_table}. 
The training time improvement of \name{} ranges from $0.7\times$ up to $16.2\times$, and the classification latency improvement ranges from $2\times$ up to $30.8\times$. 

\section{Summary and Future work}
\label{sec:future-work}

This paper presents \name{}, an efficient DTEM classification framework that augments standard DTEM classifiers to obtain lower model memory size, training time and classification latency, while obtaining competitive and often better anomaly detection than the standard (baseline) models. We also believe that \name{} can be extended to support multi-class classification. The straightforward approach is having $k$ \fg models for $k$ classes. However, an immediate concern is the scalability of this approach with respect to the number of classes. Thus, better approaches that result in less \fg models should be considered in order to maintain the competitive attributes of \name{}.

\newpage

\bibliography{reference}
\bibliographystyle{sysml2019}

\end{document}